\def\year{2022}\relax
\documentclass[letterpaper]{article} 
\usepackage{aaai22}  
\usepackage{times}  
\usepackage{helvet}  
\usepackage{courier}  
\usepackage[hyphens]{url}  
\usepackage{graphicx} 
\usepackage{algorithm,algorithmic}
\usepackage{amsmath}
\usepackage{natbib}  
\usepackage{caption} 
\DeclareCaptionStyle{ruled}{labelfont=normalfont,labelsep=colon,strut=off} 
\usepackage{algorithm}
\usepackage{algorithmic}
\usepackage{multicol}
\usepackage{multirow}
\usepackage{newfloat}
\usepackage{listings}
\floatstyle{ruled}
\newfloat{listing}{tb}{lst}{}
\floatname{listing}{Listing}
\title{Dynamic Key-value Memory Enhanced Multi-step Graph Reasoning for Knowledge-based Visual Question Answering}
\author {
    Mingxiao Li\textsuperscript{\rm 1},
    Marie-Francine Moens\textsuperscript{\rm 1}
    }
    
\affiliations {
    \textsuperscript{\rm 1} KU Leuven \\
    mingxiao.li@kuleuven.be,
    sien.moens@cs.kuleuven.be
    }

\usepackage{bibentry}

\begin{document}

\maketitle
\begin{abstract}
Knowledge-based visual question answering (VQA) is a vision-language task that requires an agent to correctly answer image-related questions using knowledge that is not presented in the given image. It is not only a more challenging task than regular VQA but also a vital step towards building a general VQA system. Most existing knowledge-based VQA systems process knowledge and image information similarly and ignore the fact that the knowledge base (KB) contains complete information about a triplet, while the extracted image information might be incomplete as the relations between two objects are missing or wrongly detected. In this paper, we propose a novel model named dynamic knowledge memory enhanced multi-step graph reasoning (DMMGR), which performs explicit and implicit reasoning over a key-value knowledge memory module and a spatial-aware image graph, respectively. Specifically, the memory module learns a dynamic knowledge representation and generates a knowledge-aware question representation at each reasoning step. Then, this representation is used to guide a graph attention operator over the spatial-aware image graph. Our model achieves new state-of-the-art accuracy on the KRVQR and FVQA datasets. We also conduct ablation experiments to prove the effectiveness of each component of the proposed model.\footnote{The code is released on: https://github.com/Mingxiao-Li/DMMGR}
\end{abstract}

\section{Introduction}
Over the past few years, the domain of visual question answering (VQA) \cite{antol2015vqa} has attracted great attention and witnessed significant progress \cite{antol2015vqa,lu2016hierarchical, hudson2019gqa}. However, most VQA models 
cannot answer questions that require external knowledge beyond what is provided in the image. Considering the top example in Figure 1, the question is "What is the relation between the object that belongs to the category of eukaryotes and the fork in the image?". To correctly answer this question, it is necessary to both understand the visible content in the image and incorporate the external knowledge that a cucumber belongs to the biological category of eukaryotes. When facing such challenging questions, we humans can easily combine the image content with general knowledge that is required for answering this question, while many current VQA models fail due to their incapability to utilize external knowledge.

To bridge this gap between human behavior and current VQA models, several knowledge-based VQA datasets have been proposed. \citet{wang2017fvqa} introduced the "fact-based VQA (FVQA)" task and developed the first KVQA dataset containing images, questions with answers and a knowledge base (KB) of fact triplets extracted from different sources including ConceptNET \cite{speer2017conceptnet}, WebChild \cite{tandon2014webchild} and DBPedia \cite{auer2007dbpedia}. Recently, \citet{2021krvqr} introduced the first large-scale knowledge-based VQA dataset: Knowledge-routed VQA (KRVQR) that contains a KB for answering the questions. In this work, we mainly focus on the KRVQR dataset and also test our model on the FVQA dataset. Other VQA datasets that require external knowledge exist \cite{marino2019okvqa,Jain2021SelectSS} but here the task is to search for external knowledge, which is not the scope of this work. 

\begin{figure}
  \begin{center}
    \includegraphics[width=0.45\textwidth]{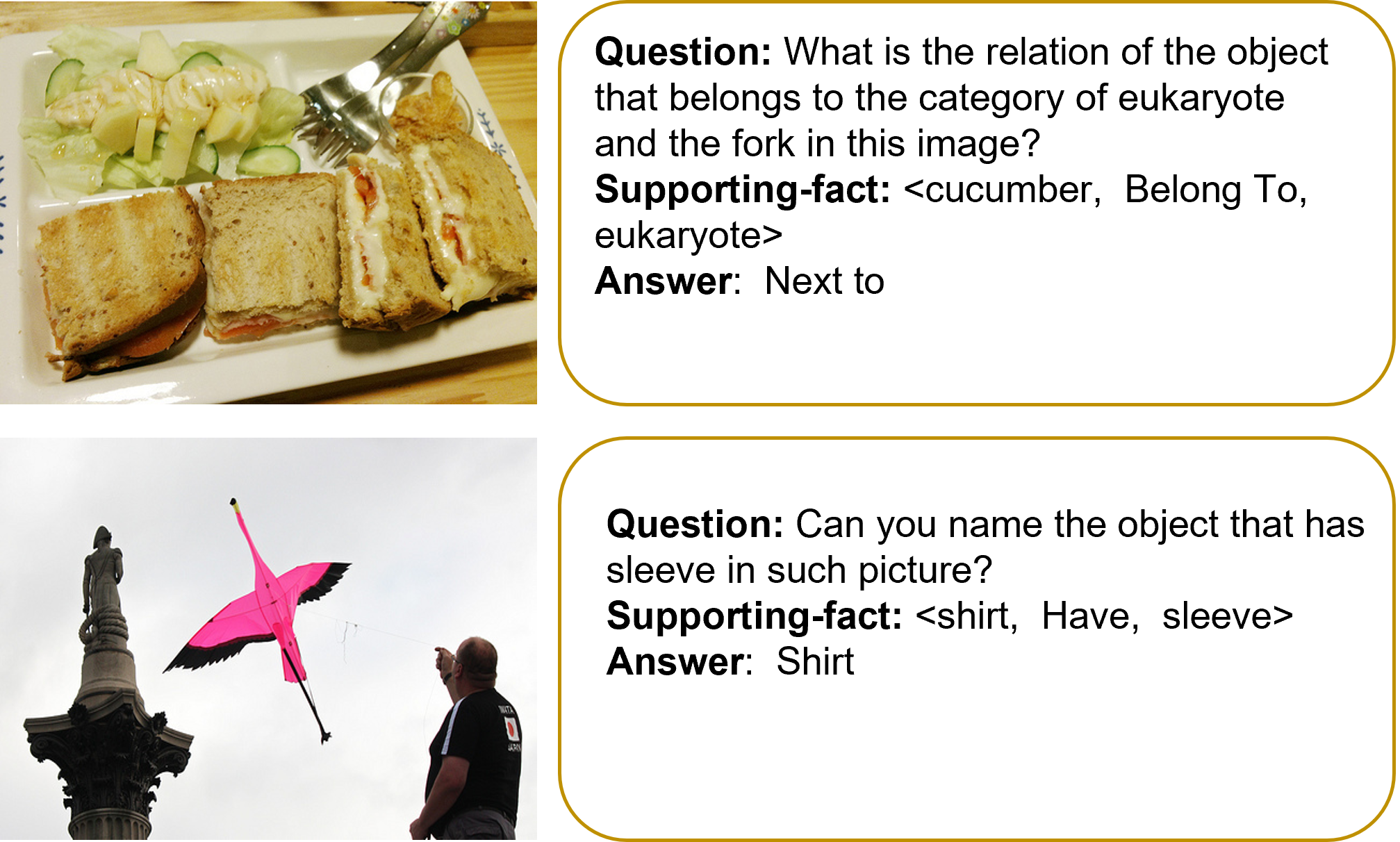}
  \end{center}
  \caption{Two examples taken from the KRVQR dataset. To correctly answer the question in the image, the model should be able to utilize supporting facts which cannot be seen in the image. 
  }
\end{figure}

Graph-based approaches \cite{narasimhan2018out,mucko} have achieved successes in the field of knowledge VQA. They use one or several graphs to represent the information sources and conduct cross-graph learning or different fusion methods to infer the answer to the given question. However, they either suffer from the drawbacks of ignoring the dynamics of multi-step reasoning, or from performing an identical reasoning procedure for both the knowledge facts and the image. Such approaches neglect that the knowledge facts provide the explicit information of a triplet, while the image graph contains only implicit information about the image.

In this work, we propose the DMMGR or \textbf{D}ynamic knowledge \textbf{M}emory enhanced \textbf{M}ulti-step \textbf{G}raph \textbf{R}easoning model, which performs explicit and implicit reasoning over a KB and a spatial-aware image graph, respectively. Specifically, we see the reasoning over the knowledge base as a problem of performing key addressing and value reading over a key-value memory and propose a novel dynamic key-value knowledge memory module to learn a question-aware knowledge representation at each reasoning step. This is different from previous key-value memory networks \cite{2016keyvalue,2019enhancingkeyvalue} whose key is the subject and relation of a triplet and value is the object.
Our proposed module dynamically learns a question representation that can reason about the subject, relation and object of a knowledge triplet. We depict the image as a spatial-aware image graph where the nodes are the embeddings of the objects detected using Faster-RCNN \cite{fasterrcnn} and the edges are embeddings of their relative positions. 
Inspired by \cite{know_scene,zareian2020bridging} who leveraged common sense knowledge for scene graph generation, we use a knowledge-aware question representation, which is learned by applying explicit reasoning over the knowledge memory. This question representation performs implicit reasoning over the spatial-aware image graph. DMMGR thus implements multi-step reasoning by iteratively performing explicit reasoning over the dynamic knowledge memory and implicit reasoning over the spatial-aware image graph.

In summary, the main contributions of this paper are as follows, (1) We propose a novel dynamic knowledge memory module that learns a representation of knowledge triplets and generates a knowledge-aware question representation. (2) We introduce a question and knowledge guided graph reasoning module, where we use the representation of related knowledge triplets to guide the reasoning over a sparse spatial-aware image graph. (3) We perform an ablation study to verify the contribution of each model component, and attention visualization shows that our model has good interpretability.  
\begin{figure*}[h]
  \vspace{-5pt}
  \centering
    \includegraphics[width=2.0\columnwidth]{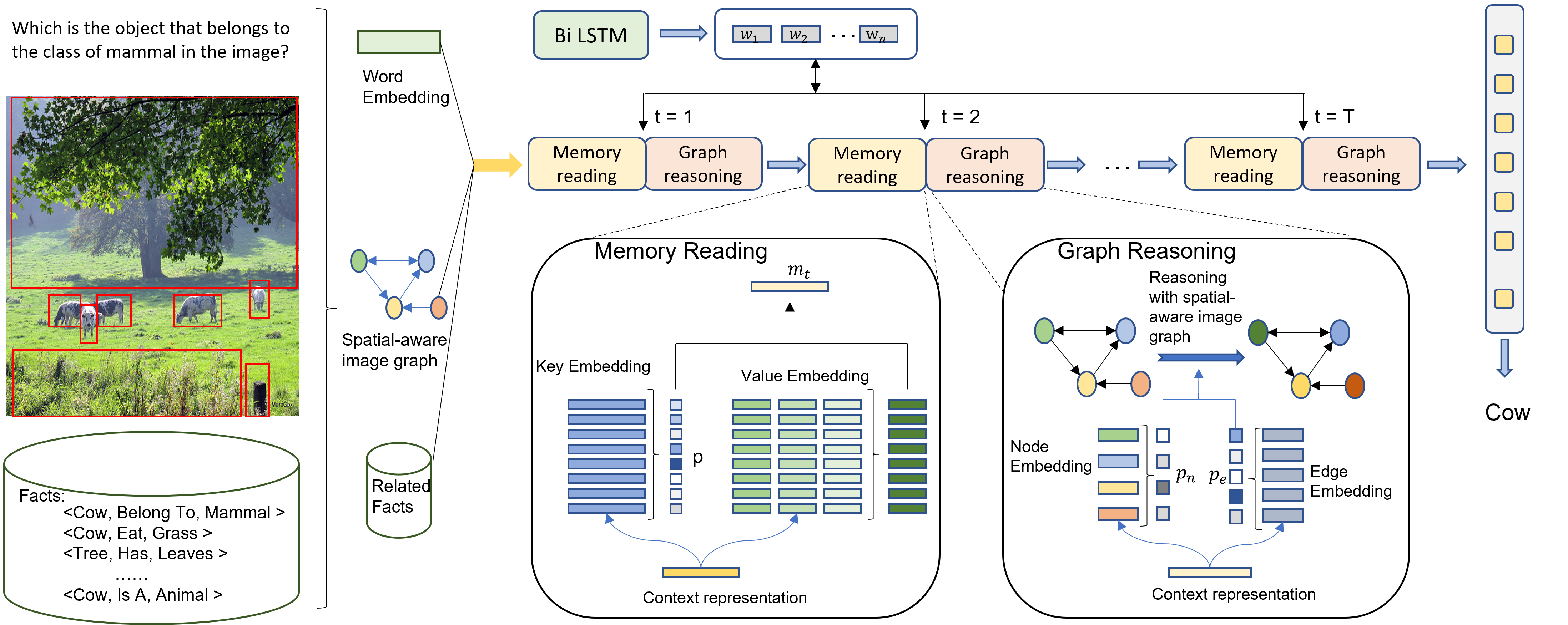}
  \vspace{-5pt}
  \caption{The architecture of our proposed DMMGR model. The module specialized in iterative reasoning consists of two components: Dynamic memory reading aimed at reading question related knowledge from the retrieved knowledge facts and resulting in a knowledge-aware question representation, and the spatial-aware image graph reasoning which performs a question and knowledge guided graph attention operation to infer the image information that is required to answer the question.}
  \vspace{-5pt}
\end{figure*}
\section{Related Work}
\subsection{Visual Question Answering}
The VQA task, where a VQA agent is expected to correctly answer a question related to an image, was proposed by \citet{antol2015vqa}. Most of the early VQA models \cite{antol2015vqa,andreas2016neural,Ben-younes_2017_ICCV,fukui2016multimodal,lu2016hierarchical,ma2016learning} integrate a CNN-RNN based architecture that fuses the RNN encoding of the question and the CNN encoding of the image to predict the answer, possibly improved by attention mechanisms to highlight the visual objects that are related to the question  \cite{yang2016stackedatt,anderson2018bottomatt,lu2016hierarchical}. Recently, graph neural networks that represent the image as a scene graph, where nodes are objects and edges are relations between two connected objects, has attracted attention in many vision-language tasks including VQA. \citet{teney2017graph} represented both a language question and an image as two graphs and applied graph attention and an aggregation operator to infer the corresponding answer. \citet{hu2019language} and \citet{wang2019neighbourhood} used question guided image graph attention to generate a question-aware image graph representation. Different from the aforementioned graph models, \citet{norcliffe2018learning} proposed a graph learner, where a fully connected image graph whose nodes are region features, and edges represent position information was constructed based on the given question, and this model performs graph convolution and max pooling to predict the answer.

\subsection{Knowledge-Based VQA}
Knowledge-based VQA (KVQA) requires the model to use knowledge outside the image to answer questions correctly. Compared to the original VQA task, KVQA is relatively less explored (e.g. 
\citet{wang2017fvqa, marino2019okvqa}) 
\citet{narasimhan2018out} have first introduced a graph based approach to KVQA and apply a graph convolution to the fact graph to infer the answer, while \citet{ziaeefard2020towards} use graph attention and multimodal fusion to reason over both the image graph and fact graph. Another work \cite{mucko} depicts an image as three graphs: a semantic graph built on the results of dense captioning of image regions, a fact graph representing relevant knowledge triplets, and a fully connected image graph where nodes represent region features, and iteratively perform question guided inter- and intra-graph attention to answer the question. In contrast, we use a dynamic key-value memory to represent the triplets and use a sparse spatial-aware image graph whose nodes are the object category embeddings, and edges are the relative position embeddings of two objects. As an extension of our proposed model, we have evaluated the integration of a semantic graph based on dense captioning of the image regions.

\subsection{Key-value Memory Networks}
A key-value memory network \cite{2016keyvalue} is seen as an extension of a memory network, \cite{weston2014memory,sukhbaatar2015end}. However, different from memory networks, key-value memory networks save context as key-value pairs and split the reasoning process into key addressing and value reading. Key-value memory networks have been widely used in knowledge triplet-based question answering \cite{2016keyvalue,2019enhancingkeyvalue}. For a knowledge triplet $<subject, relation,object>$, a key-value memory network saves the subject and relation as a key and the object as a value, which restricts its usage. In this paper, we introduce a dynamic key-value memory module whose key is composed of all the information of a triplet and whose value is a question-aware triplet representation. Such a memory structure is flexible enough so that it can reason about not only the object but also the subject and relation of a knowledge triplet.

\section{Methodology}
Given a question $Q$, an image $I$ and a KB $K$ that consists of a set of facts 
${f_1,f_2,...,f_n}$, a KVQA 
model aims to predict the answer $A$ to $Q$ 
by reasoning over the image and the KB. Each fact is represented as a resource description framework (RDF) triplet of the form $f_i=(e_1,r,e_2)$, where $e_1$, $e_2$ are entities and $r$ is the relation between $e_1$ and $e_2$. In our model, we first conduct explicit reasoning over the dynamic key-value memory module to extract question related knowledge information. Then we use both extracted knowledge information and the question representation to perform implicit reasoning over the spatial-aware image graph. The complete model is constructed by stacking both the explicit key-value memory reading module and the implicit spatial-aware image graph reasoning module. Figure 2 shows the details of our proposed model.

\subsection{Key-Value Memory Construction}

\textbf{Fact Retrieval.}  To retrieve the relevant facts $f_{rel}$ from the KB, we detect the nouns mentioned in the question and the objects in the image using the Stanza natural language processing (NLP) tool \cite{qi2020stanza} and the pretrained Faster-RCNN with RestNet-101 backbone 
\cite{fasterrcnn}, respectively. We then use a pretrained GloVe 
word embedding \cite{glove} to represent each entity and relation in the knowledge triplet, the nouns in the question and the detected image objects, and sort the triplets based on the average cosine similarity between every word in a triplet and the nouns in the question and detected image objects ignoring pairs with zero average similarity. 
Finally, the top $k=5$ facts with the highest average cosine similarity values are retained for predicting the answer to the question.

\noindent\textbf{Memory Construction.} A factual triplet consists of 
$<$ subject, relation, object $>$, such as $<$ mouse, related to, keyboard $>$. We store the extracted triplets in a key-value memory structure. This is different from previous works \cite{2016keyvalue,2019enhancingkeyvalue} that take the subject and relation as a key and the object as a value, which strongly limits their usage in reasoning. We use the average embedding of subject, relation and object as key, and the value contains each element of a triplet: $\{F_f: [F_s,F_r,F_o]\}$, where $F_f$ is the mean GloVe embedding of the words that form an element of the triplet, and $F_s$, $F_r$, $F_o$ are the embeddings of the subject, relation and object, respectively. Our proposed key-value structure is capable of providing all information including subject, relation and object of a triplet. 

\subsection{Visual Spatial-aware Graph Construction}
Given an image $I$, we detect objects in the image and keep the top $r=36$ detected objects $O = \{o_i\}_{i=1}^r$. Each object is associated with a label representation $v_i \in R^{d_v} (d_v = 300)$ which is the average GloVe embedding \cite{glove} of the predicted object category, and a spatial vector $b_i \in R^{d_b} (d_b=4)$ consisting of the coordinates of the top-left and bottom-right corners of the bounding box. The previous work \cite{mucko,hu2019language} use a fully connected spatial-aware image graph, which not only increases the computational cost but also introduces noisy relations between objects, we construct a sparse and $spatial-aware$ image graph $G^V=\{V^V,E^V\}$ over the objects $O$ by connecting only one object with its top-$5$ nearest neighbors. The distance between two objects is the squared distance between the centers of their two bounding boxes. Figure 3 shows an example of how we create this spatial-aware image graph. Each node in the node set $V^V = \{v_i^V\}_{i=1}^M$ corresponds to a detected object $o_i$ represented by its label representation, and the edges $e_{ij}^V\in E^V$ denote the relations between pairs of objects. We use a 5-dimensional relative spatial vector $r_{ij}^V=[\frac{x_{i}^{c}-x_{j}^{c}}{\sqrt{w_ih_i}},\frac{y_i^c-y_j^c}{\sqrt{w_ih_i}},\frac{w_j}{w_i},\frac{h_j}{h_i},\frac{w_jh_j}{w_ih_i}]$, where $x_i^c,y_i^c$ are the center coordinates and $w_i,h_i$ are the width and height of the bounding box of object $i$, respectively, to encode edge features in the graph. 

\begin{figure}[h]
  \vspace{-10pt}
  \begin{center}
    \includegraphics[width=0.45\textwidth]{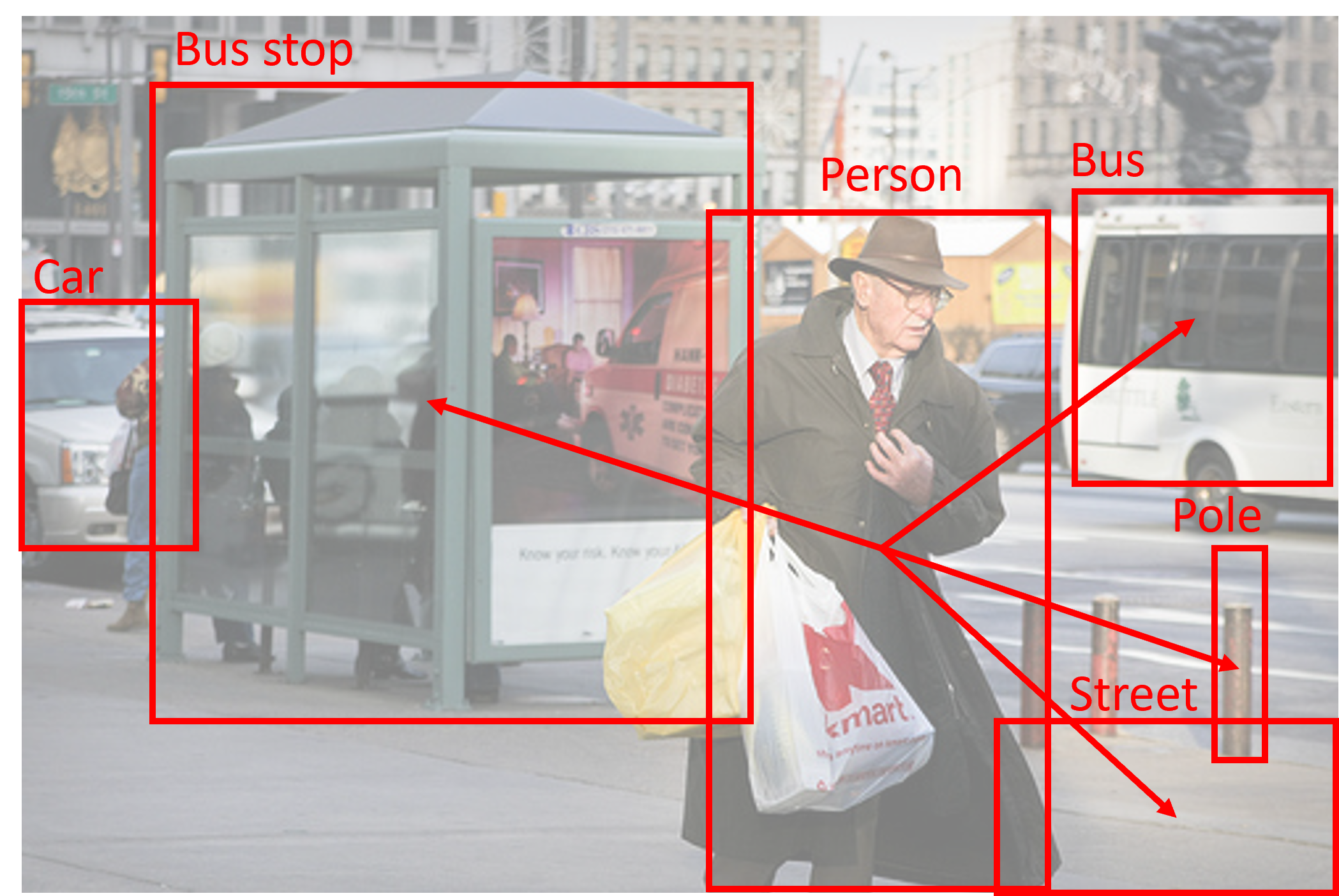}
  \end{center}
  \vspace{-10pt}
  \caption{An example of a spatial-aware image graph. We only show the connections between person and its neighbours. There is no connection between person and car as the car is far from the person.}
  \vspace{-10pt}
\end{figure}

\subsection{Iterative Reasoning Module}
\label{section:1}
Algorithm 1 illustrates the flow of our proposed iterative reasoning module.
The $;$ in algorithm 1 denotes the concatenation operation, and $T$ is the total number of reasoning steps. The details of each module are introduced below.

\noindent\textbf{Question Processing. } To enhance the model's ability to perform multi-step reasoning, we generate a 
question vector for each reasoning step $t$ (where $t=1,\cdots,T$). Specifically, the given question $Q$ of length $S$ 
is converted into a sequence of GloVe word embeddings that are further processed by a 2-layer bi-directional long short-term memory (LSTM) module to generate a sequence of contextually aware word representations:
\begin{equation}
    [h_1,h_2,\cdots,h_S] = BiLSTM(Q) \label{1} \tag{1}
\end{equation}
where $h_s=[\overrightarrow{h_s};\overleftarrow{h_s}]$ is the concatenation of the forward and backward hidden state of words at position $s$ from the last layer of the bi-directional LSTM output. At each reasoning step $t$, context vector $c^t$ attends over the sequence of question words, and the question representation $q^t$ is obtained as follows:  
\begin{align*}
    \alpha_{s}^t &=softmax(W_1(h_s\odot (W_2^tReLU(W_3c^t))) \label{2}\tag{2}  \\
    q^t &= \sum_{s=1}^S\alpha_s^t h_s \label{3} \tag{3}  
\end{align*}
where $W_1$, $W_2^t$ (as well as $W_3$,$\cdots$, $\omega_v^t$, $\omega_e^t$ mentioned below) are learned parameters, and $\odot$ denotes element-wise multiplication. Parameters with $t$ superscript are learned separately for each iteration, while those without $t$ superscript are shared across iterations. The $q^t$ can be seen as the reasoning step-aware question representation at step $t$. In the first step, the reasoning step-aware vector: $c^1=[\overrightarrow{h}_S;\overleftarrow{h}_1]$, which is the concatenation of the final hidden states obtained from the forward and backward LSTM passes. From the second step to the $T$th step, $c^t$ is evaluated as:
\begin{align*}
    c^t = W_4^t(ReLU(W_5[R^{t-1};I^{t-1}])) \label{4} \tag{4} 
\end{align*}
where $R^{t-1}$, $I^{t-1}$ are the knowledge-aware question representation and visual representation at step $t-1$, respectively, which are explained in details below. 

\noindent\textbf{Key Addressing and Value Reading.} Key addressing is a process that involves finding the most relevant knowledge triplet for a given question. Following previous work \cite{2019enhancingkeyvalue}, we compute the relevance probability $p_i$ between the question representation $q$ and each key representation $k$ as: 
\begin{align*}
    \hat{q} &= ReLU(W_6(ReLU(W_7q) \label{5} \tag{5} \\
    \hat{k}_i &= ReLU(W_8(ReLU(W_9k_i) \label{6} \tag{6} \\
    p_i &= softmax(\hat{q}\cdot\hat{k}_i^T) \label{7} \tag{7}
\end{align*}
As each value of our key-value memory module contains all three elements (subject, relation and object) of a triplet, we further apply an attention mechanism to compute the question-guided value embedding $\hat{t}_i$ for each value:  
\begin{align*}
    \hat{t}_{ij} &= ReLU(W_{10}(ReLU(W_{11}t_{ij})) \label{8} \tag{8} \\
     s_{ij} &=  (1-softmax(\hat{q}\cdot\hat{t}_{ij}^T))/2.0 \label{9} \tag{9} \\
     \hat{t_i} &= \sum_{j=1}^J s_{ij}\hat{t}_{ij} \label{10} \tag{10}
\end{align*}
where $J=3$ is the number of elements of a triplet; $s_{ij}$ is the attention probability of each component in a knowledge triplet, and the divisor $2.0$ is a normalization factor (normalize the sum of $s_{ij}$ to one)\footnote{
$\sum_{j=1}^3(1-s_{i,j}) = 3-\sum_{j}(z_{ij}) = 2$ where $z_i= softmax(\hat{q}\hat{t}^T_{i})$.}
The value of memory $m^t$ is then computed by taking the weighted sum over the question-guided value embeddings using the relevance probabilities; this value is further used to reason over the spatial-aware image graph.
\begin{align*}
    m^t=\sum_{i=1}^Kp_i\hat{t}_i \label{11}\tag{11}
\end{align*}
where $K$ is the number of knowledge triplets stored in the key-value memory module.

\noindent\textbf{Question and Knowledge Guided Node Attention.} We first merge the question information and the output of the key-value memory module to generate a knowledge-aware question representation. 
\begin{align*}
    R^t = W_{11}^t (ELU (W_{12} [q^t;m^t] ) ) \label{12}\tag{12}
\end{align*}
where $q^t$ and $m^t$ are the question and memory embeddings generated during the $t-th$ reasoning iteration, respectively, and $ELU$ is the exponential linear unit activation function used here to avoid the dead neuron problem. We then use an attention mechanism to compute the relevance of each node of the spatial-aware image graph corresponding to the knowledge-aware question representation. The relevance scores are evaluated as:
\begin{align*}
    \alpha_i = softmax(\omega_v tanh(W_{13}v_i+W_{14}R^t)) \label{13}\tag{13}
\end{align*}
where $v_i$ is the node representation of the spatial-aware image graph.

\noindent\textbf{Question and Knowledge Guided Edge Attention} We apply the same attention mechanism to compute the importance between edge $e_{ij}$ and the knowledge-aware question representation $R^t$, which is evaluated as: 
\begin{align*}
    \beta_{ij} = softmax(\omega_e tanh(W_{15}e_{ij}+W_{16}R^t) \label{14}\tag{14}
\end{align*}

\noindent\textbf{Multi-head Spatial-aware Image Graph Attention.} Based on the node and edge attention weights computed in Eq. \ref{13} and Eq. \ref{14}, respectively, the node representations of the spatial-aware image graph are updated with multi-head graph attention \cite{graphatt}. 
\begin{align*}
 m_i^k &= \sum_{j\in \mathcal{N}_i} ([\alpha_jW_{17}v_j; \beta_{ij}W_{18}e_{ij}])\label{15} \tag{15}\\
 h_i^k &= \alpha_iReLU(W_{19}[m_i^h; W_{20}v_i]) \label{16} \tag{16} \\
 \hat{v}_i &= LayerNorm(ELU(W_{21}[h_i^1;h_i^2;\cdots;h_i^H]\label{17} \tag{17}))
\end{align*}
where $\mathcal{N}_i$ is the neighborhood set of node $v_i$ and $H$ is the number of heads. Once all nodes are updated, the max pooling operation is conducted over all nodes to obtain the visual representation at reasoning step $t$:
\begin{align*}
    I^t = MaxPooling(\{v_i\}_{i_=1}^M) \label{18}\tag{18}
\end{align*}

\noindent\textbf{Final Prediction.} We iteratively perform knowledge key addressing, 
value reading and spatial-aware image graph attention reasoning for $T$ steps. At step $T$ the given knowledge-aware question representation $R^T$ and visual representation $I^T$ are concatenated and processed by a 2-layer linear transformation to predict the answer to the question. During training, we simply use the cross-entropy loss function (see equation below), to optimize the differences between the predicted answer and ground truth answer.
\begin{align}
    L = -\frac{1}{N}\sum_{i=1}^N\sum_{c=1}^Ly_{c}log(\hat{y}_c) \label{19} \tag{19}
\end{align}
where $N$ and $L$ are the number of training samples and candidate answers, respectively. $y_{c}$ is the ground truth answer, and $\hat{y}_c$ is the predicted answer.

\section{Experiments}
\textbf{Datasets}. In this paper we mainly focus on the KRVQR \cite{2021krvqr} dataset and also test our model on the FVQA \cite{wang2017fvqa} dataset, as these are the only two knowledge VQA datasets that provide the knowledge base for answering the questions. The KRVQR dataset consists of 32910 images and 157201 question answer pairs, divided into training, validation and test sets with proportions of $60\%$, $20\%$ and $20\%$, respectively. The FVQA dataset contains $2190$ images and $5826$ questions, which are further split into training ($2927$) and test ($2899$) sets. The KRVQR dataset contains $43.5\%$ one-step reasoning and $56.5\%$ two-step reasoning questions, and the FVQA dataset contains only one-step reasoning questions \cite{wang2017fvqa}. Two-step reasoning questions need to reason over two relations to infer the answer, while one-step reasoning questions need only one relation, where relations can be found in the KB and/or image.  
An example of a two-step and an example of a one-step reasoning question are presented on the top and bottom of Figure 1, respectively. 


\textbf{Evaluation Metrics}. Following the literature \cite{2021krvqr, wang2017fvqa}, we evaluate our model using top-1 accuracy (KRVQR and FVQA datasets) and top-3 accuracy (FVQA dataset).

\textbf{Implementation Details}. 
We implement our model using the PyTorch framework \cite{paszke2019pytorch}. The hidden size of the LSTM encoder is set to 512, and the dropout rate is 0.1. We set the sizes of the dynamic key-value memory embeddings and graph node and edge embeddings to 300 and 1024, respectively. The number of reasoning steps\ref{section:1} is set to $2$ as the questions in the dataset require maximum 2 reasoning steps. All these parameters are selected based on the validation results. The model is trained using the Adam algorithm \cite{kingma2014adam} with a base learning rate of $1e^{-4}$. We 
gradually increase the learning rate over the first two epochs and start decaying the learning rate at epoch 20. The best model is trained for approximately 40 epochs with a batch size of 128. 

\begin{algorithm}[h]
\renewcommand{\algorithmicrequire}{\textbf{Input:}}
\renewcommand{\algorithmicensure}{\textbf{Output:}}
\caption{Iterative Reasoning Module}
\begin{algorithmic}[1]
\REQUIRE ~~\\ Question $Q$, key-value memory $M$, spatial-aware image graph $G$ 
\ENSURE ~~\\ Answer prediction $P$
\STATE Process $Q$ based on Equation \ref{1}
\STATE Initialize $c^1=[\overrightarrow{h_S};\overleftarrow{h}_1]$
\FOR{$t=1$ to $T$}
\STATE Obtain $q^t$ based on Equations \ref{2} and \ref{3}
\STATE Perform key addressing and value reading to obtain the representation of the knowledge triplets $m^t$ based on Equations \ref{5}$\sim$\ref{11}
\STATE Obtain the knowledge-aware question representation $R^t$ based on Equation \ref{12}
\STATE Compute the question knowledge-guided visual node and edge attention based on Equations \ref{13} and \ref{14}
\STATE Update the visual node representation $v_i$ based on Equations \ref{15}$\sim$\ref{17}
\STATE Obtain the spatial-aware image graph output $I^t$ based on Equation \ref{18}
\IF{$t<T-1$}
\STATE Update context representation $c^t$ based on Equation \ref{4}
\ENDIF
\ENDFOR
\STATE $P = Linear(Linear([R^T;I^T]))$
\RETURN $P$
\end{algorithmic}
\end{algorithm}

\section{Results}
Table 1 illustrates the accuracy comparison between our DMMGR model with other models including the state-of-the-art VQA and knowledge based VQA models. The Q-type \cite{2021krvqr}, LSTM \cite{2021krvqr}, FiLM \cite{perez2018film}, MFH \cite{yu2018beyond}, UpDown \cite{anderson2018bottomatt}, and state-of-the-art VQA model MCAN \cite{mcan} were reimplemented and tested on the KRVQR dataset in the work of \citet{2021krvqr}, and we copy the obtained results here. As running software is not publicly available, we reimplement the Mucko model and test it on the KRVQR dataset. All our results are average over $5$ runs. As presented in Table 1, our DMMGR model significantly outperforms all the other models and surpasses the current state-of-the-art model (KM-net) by approxiFmately $\textbf{6\%}$ in terms of accuracy. We also present the results of DMMGR extended with a semantic graph based on dense captioning of image regions that aims at including attributes of the image regions (DMMGR + Dense Captioning) following the method described in \citet{yu2020cross}. We see a slight accuracy improvement when including the dense captioning information in the DMMGR model, however, statistical significance testing (t-test $p=0.051$) shows that no significant differences can be observed. We report the results on the FVQA dataset in Table 2. GRUC \cite{yu2020cross} is the current state-of-the-art model for the FVQA dataset. The complete GRUC model also integrates information from a pretrained image dense caption generation model. 
The table shows that both the DMMGR and DMMGR + Dense Captioning models outperform GRUC, thus obtaining new state-of-the-art performance for the FVQA dataset. 
Integrating the dense captioning information in the DMMGR model could significantly improve its accuracy on the FVQA dataset by around $2.6\%$ (t-test $p$ = $0.02$), which is different from what we observe in the KRVQR dataset. This difference might come from the difference of data distribution and types of questions in the two datasets.

\begin{table}[h]
    \centering
    \begin{tabular}{l|cc}
    \hline Model & Accuracy   \\
    \hline
      Q-type \cite{2021krvqr}   &  $8.12$ \\
      LSTM \cite{2021krvqr}  &  $8.81$ \\
      FiLM \cite{perez2018film} & $16.89$ \\
      MFH \cite{yu2018beyond}  & $19.55$ \\
      UpDown \cite{anderson2018bottomatt} &  $21.85$ \\
      MCAN  \cite{mcan} & $22.23$ \\
      Mucko \cite{mucko} & $24.00$ \\
      KM-net \cite{cao2019explainable} & $25.19$ \\
     \hline 
     \textbf{DMMGR} (2-steps)& $31.4$ \\ 
     \textbf{DMMGR + Dense Captioning (2-steps)} & $31.8$ \\
     \hline
    \end{tabular}
    \caption{Top-1 accuracy comparisons among different models on the KRVQR dataset.}
    \label{tab:my_label}
\end{table}

\begin{table}[h]
    \centering
    \begin{tabular}{l|cc}
    \hline 
    \multirow{2}*{Model} & \multicolumn{2}{c}{Accuracy} \\
     ~ & top-1 & top-3  \\
    \hline 
    FVQA (Ensemble) \cite{wang2017fvqa}   &  $58.76$ \\   
    $STTF^1$ & $62.20$ & $75.60$ \\
    $OB^2$  & $69.35$  & $80.25$\\ %
    Mucko \cite{mucko}  & $73.06$  & $85.94$\\ 
    GRUC \cite{yu2020cross}  & $79.63$ & $91.20$ \\  
    GRUC (without Semantic graph) & $78.05$ & $87.70$ \\
    \hline 
    \textbf{DMMGR (1-step)} & $78.6$ &  $90.6$ \\
    \textbf{DMMGR + Dense Captioning (1-step)} & $81.20$ & $95.38$  \\
    \hline 
    \end{tabular}
    \caption{Top-1 and top-3 accuracy of the different models obtained on the FVQA dataset.(1:\citet{narasimhan2018straight},2:\citet{narasimhan2018out})}
    \label{tab:my_label}
\end{table}
\section{Ablation Study}
To verify the effectiveness of each component of our model, we conduct an extensive ablation study of the results obtained on the KRVQR dataset, which is the most challenging dataset (Table 3, 4 and 5). They reveal the impact of the iterative reasoning module, the dynamic key-value memory module and the knowledge-guided graph reasoning.

\subsection{Does the number of reasoning steps matter?}
We first verify the contribution of the iterative reasoning module by performing experiments using the DMMGR model with different numbers of reasoning steps. From Table 3, we can observe that the DMMGR model with two reasoning steps has the best performance, which is slightly higher than that of the DMMGR model with only one reasoning step. However, having more than two reasoning steps, such as three and four, dramatically decreases the accuracy by more than $4\%$. This is not surprising, as the KRVQR dataset only contains one-step reasoning $(43.5\%)$ and two-step $(56.5\%)$ reasoning questions. 

\begin{table}[h]
    \centering
    \begin{tabular}{l|cc}
    \hline Model (DMMGR) & Accuracy \\
    \hline 
     $1$ step  & $30.5$  \\
     $2$ steps & $31.4$ \\
     $3$ steps & $27.1$ \\
     $4$ steps & $26.2$ \\
    \hline
    \end{tabular}
    \caption{Results in terms of top-1 accuracy of the DMMGR model obtained on the KRVQR dataset considering different numbers of reasoning steps.}
    \label{tab:my_label}
\end{table}

\begin{figure*}[h]
  \vspace{-10pt}
  \centering
    \includegraphics[width=1.7\columnwidth]{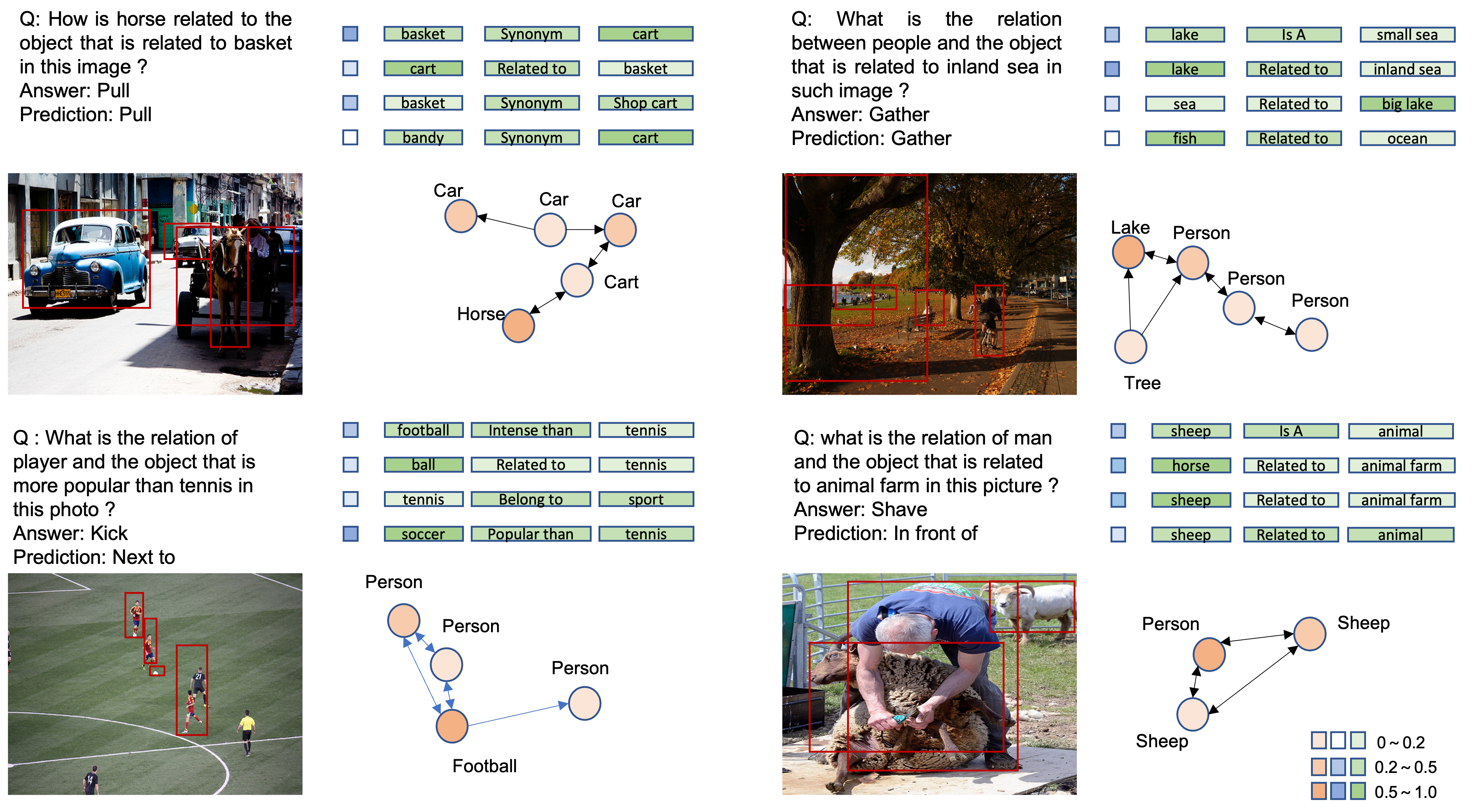}
  \vspace{-10pt}
  \caption{Attention visualization of the DMMGR model. The top and bottom maps are the attention weights of randomly selected samples from the test set that are correctly and incorrectly answered by the model, respectively. Different colors are used to represent the attention weights in the spatial-aware image graph and knowledge memory module, where deeper colors denote higher attention weights." Answer" represents the true answer and "Prediction" stands for the output of the model.(More examples can be found in Figure 5)}
  \vspace{-10pt}
\end{figure*}
\begin{figure*}
  \centering
    \includegraphics[width=1.8\columnwidth]{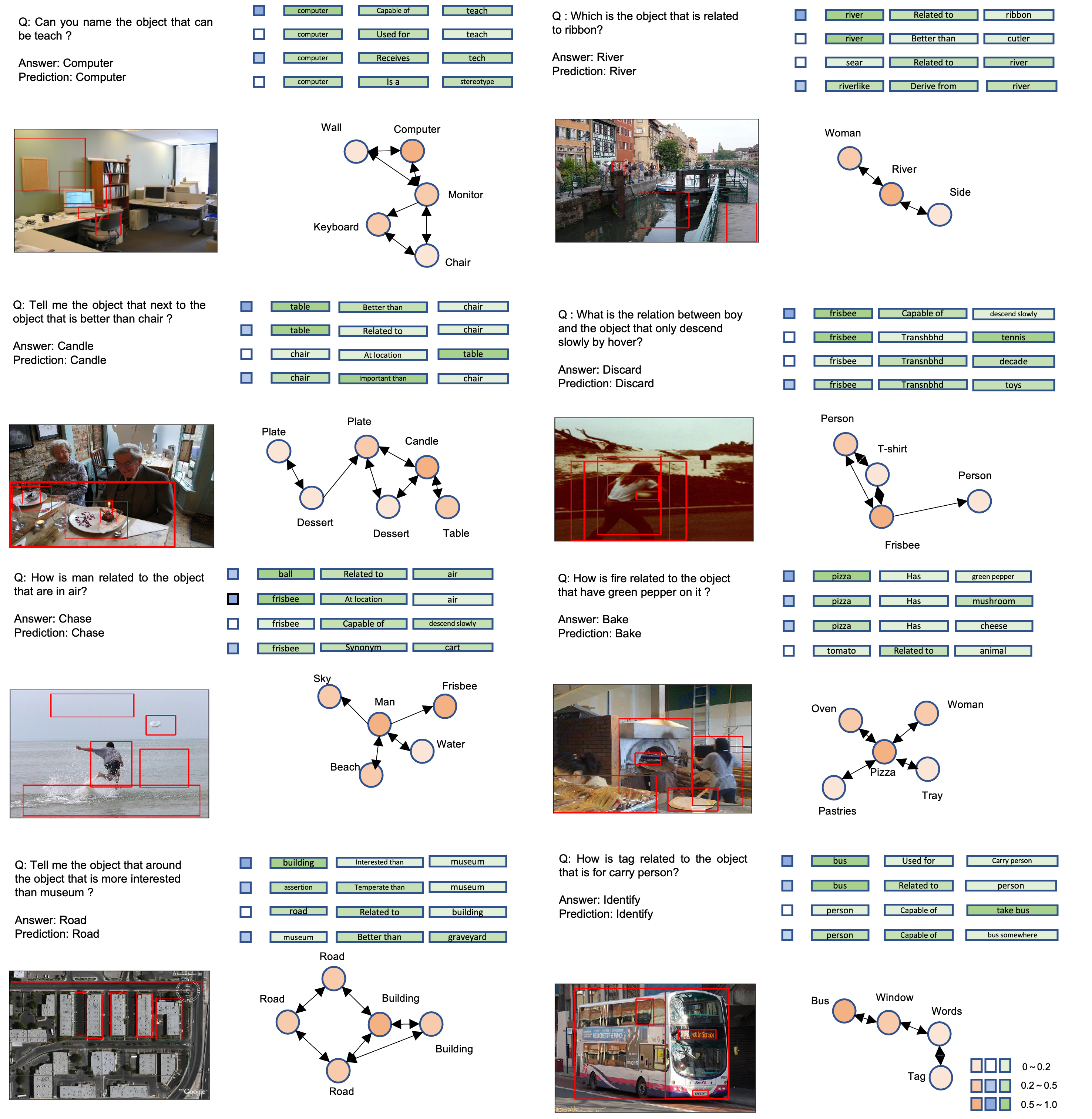}
  \vspace{-10pt}
  \caption{Attention visualization of the DMMGR model. All the samples are randomly selected from the test set of the KRVQR dataset. Different colors are used to represent the attention weights in the spatial-aware image graph and knowledge memory module, where deeper colors denotes higher attention weights." Answer" represents the true answer and "Prediction" stands for the output of the model.}
  \vspace{-10pt}
\end{figure*}

\subsection{Does the Proposed Key-value Memory Really Help?}
We next study the advantage of using a dynamic key-value memory module (Table 4). The experiments are conducted by replacing the memory module of the DMMGR model with different types of memory modules and testing the new models on the KRVQR dataset. The first model in Table 4 integrates a simple memory module where each slot is the average embedding of a knowledge triplet. The second model in Table 4 has the regular key-value memory module whose keys are the representation of subject and relation, and values are the object representation of a knowledge triplet \cite{2016keyvalue}. The third model is the dynamic key-value memory module proposed in this paper. The results show that the accuracy of our proposed memory model surpasses the other two models by approximately $5\%$, which indicates that the dynamic key-value memory module has a beneficial effect on knowledge triplet reasoning. It is also rational to observe that a simple memory module slightly performs better than the common 
key-value memory module \cite{2016keyvalue}, since the latter can only infer the object of a triplet while questions in the KRVQR dataset could be related to the subject or relation of a triplet.

\begin{table}[h]
    \centering
    \begin{tabular}{c|c}
    \hline  Memory model & Top-1 accuracy \\ 
    \hline 
     Average embedding memory module & $27.3$  \\
    \hline 
     Key-value memory module & $26.5$ \\
    \hline 
     Proposed memory model & $31.4$\\
    \hline 
    \end{tabular}
    \caption{Results in terms of top-1 accuracy of the DMMGR model obtained on the KRVQR dataset considering
    different types of memory modules. 
    }
    \label{tab:my_label}
\end{table}
\subsection{Can External Knowledge Help the Model to Understand the Image Better?}
Finally, we explore whether the retrieved knowledge triplets can help the model to better understand the image. In Table 5, we compare the performance of the models with and without the use of the knowledge triplets in the reasoning module operating on the spatial-aware image graph. One can observe that the accuracy decreases by approximately $2\%$ when the reasoning module has no access to the knowledge triplets, which supports our proposed knowledge-guided image graph reasoning module. 

\begin{table}[h]
    \centering
    \begin{tabular}{c|c}
    \hline Model (DMMGR) & Accuracy \\ 
    \hline 
     w/o knowledge-guided reasoning  & $29.9$ \\ 
    \hline 
    full model & $31.4$  \\
    \hline 
    \end{tabular}
    \caption{The accuracy values obtained with and without the us of the knowledge triplets in the image graph reasoning module obtained on the KRVQR dataset.}
    \label{tab:my_label}
\end{table}

\section{Qualitative Analysis}
To further study the working mechanism of the DMMGR model, we randomly select two samples from the test set that are correctly answered by our model, and present the corresponding attention weights on top of Figure 4 (more examples are found in Figure 5).
Note, that to obtain a clear visualization map, we plot only the attention weights of some highly relevant knowledge triplets and objects in the image. The attention weights show that DMMGR not only correctly selects the most relevant knowledge triplets and addresses the correct elements, but also focuses on the related objects in the image. The bottom attention weight maps of Figure 4 are examples where the DMMGR model fails. We observe that DMMGR here fails when there is ambiguity present in the image. For example, there are more than two persons in the scene, although the model could find the most relevant information from the knowledge base, it focuses on the wrong person, which leads to an incorrect answer. Improving the model's ability of handling such confusion could be an interesting future work.
\section{Conclusion}
We have proposed a multi-step graph reasoning model that is enhanced by a novel dynamic memory, which iteratively performs explicit and implicit reasoning over a key-value triplet memory and a spatial-aware image graph, respectively, to infer the answer in a KVQA task. Our model achieves new state-of-the-art performance on both the KRVQR and FVQA datasets.

\section{Acknowledgement}
This research received funding from the Flanders AI Impuls Programme - FLAIR and from the European Research Council Advanced Grant 788506.

\bibliography{aaai22}
\appendix
\end{document}